\documentclass[letterpaper, 10 pt, conference]{ieeeconf}
\usepackage{amsmath,amsfonts}
\usepackage{algorithmic}
\usepackage{algorithm}
\usepackage{array}
\usepackage{cite}
\usepackage[caption=false,font=normalsize,labelfont=sf,textfont=sf]{subfig}
\usepackage{textcomp}
\usepackage{stfloats}
\usepackage{url}
\usepackage{verbatim}
\usepackage{adjustbox}
\usepackage{graphicx}
\usepackage[dvipsnames,table,xcdraw]{xcolor}
\usepackage{booktabs}
\usepackage{gensymb}
\usepackage{multirow}
\usepackage{hyperref}
\usepackage{arydshln}
\usepackage{svg}
\usepackage{adjustbox}
\usepackage{textcomp}
\usepackage{gensymb}
\usepackage{cleveref}

\usepackage{caption}
\usepackage{makecell}
\definecolor{synthetic}{rgb}{0.0,0.0,0.8}
\definecolor{real}{rgb}{0.6,0.0,0.1}

\definecolor{darkgreen}{rgb}{0.09, 0.45, 0.27}
\definecolor{tablecolor2}{RGB}{232, 234, 246}
\definecolor{tablecolor3}{RGB}{233, 255, 230}
\newcommand{\ddbf}[1]
{\cellcolor{tablecolor3}$\mathbf{#1}$}
\newcommand{\dd}[1]
{\cellcolor{tablecolor2}$#1$}
\newcommand{\ex}{\textcolor{black}}

\newcommand\freefootnote[1]{%
  \let\thefootnote\relax%
  \footnotetext{#1}%
  \let\thefootnote\svthefootnote%
}

\title{\LARGE \bf
ViTaS: Visual Tactile Soft Fusion Contrastive Learning for Visuomotor Learning
}
\author{
Yufeng Tian$^{*\,1,2}$,
Shuiqi Cheng$^{*\,4}$,
Tianming Wei$^{1,3}$,
Tianxing Zhou$^{3}$,\\
Yuanhang Zhang$^{5}$,
Zixian Liu$^{3}$,
Qianwei Han$^{1}$,
Zhecheng Yuan$^{1,3}$,
Huazhe Xu$^{1,3}$
}

\begin{document}

\twocolumn[{
    \begin{@twocolumnfalse}
    \maketitle
    \begin{center}
        \includegraphics[width=\textwidth]{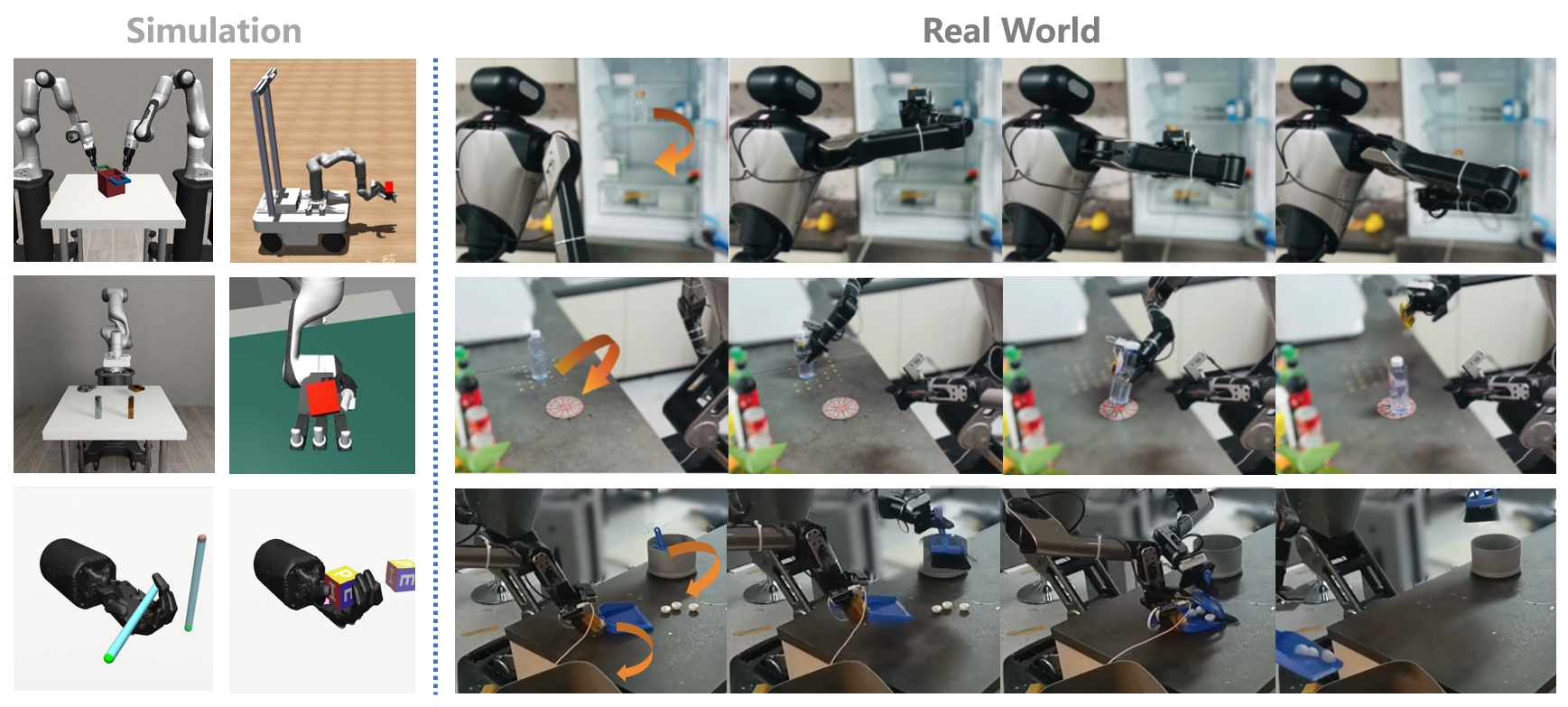}
        \captionof{figure}{\textbf{ViTaS} is capable of handling various simulation and real-world manipulation tasks including transparent objects and self-occluded scenarios, by fusing visual and tactile features for effective policy learning. }
        \label{fig:teaser}
    \end{center}
    \end{@twocolumnfalse}
}]

\begin{abstract}
Tactile information plays a crucial role in human manipulation tasks and has recently garnered increasing attention in robotic manipulation. However, existing approaches mostly focus on the alignment of visual and tactile features and the integration mechanism tends to be direct concatenation. Consequently, they struggle to effectively cope with occluded scenarios due to neglecting the inherent complementary nature of both modalities and the alignment may not be exploited enough, limiting the potential of their real-world deployment. In this paper, we present \textbf{ViTaS}, a simple yet effective framework that incorporates both visual and tactile information to guide the behavior of an agent. We introduce \textit{Soft Fusion Contrastive Learning}, an advanced version of conventional contrastive learning method and a CVAE module to utilize the alignment and complementarity within visuo-tactile representations. We demonstrate the effectiveness of our method in $\mathbf{12}$ simulated and $\mathbf{3}$ real-world environments, and our experiments show that ViTaS significantly outperforms existing baselines. Project page: https://skyrainwind.github.io/ViTaS/index.html.
\end{abstract}

\freefootnote{${}^*\,$Equal Contribution.}
\freefootnote{$^1\,$Shanghai Qi Zhi Institute. $^2\,$Harbin Institute of Technology. $^3\,$Tsinghua University. $^4$\,The University of Hong Kong. $^5$\,Carnegie Mellon University.}

\section{Introduction}


Humans are adept at performing complex manipulation tasks, such as spinning an object or cleaning a table. While vision plays a critical role, other modalities, particularly touch, also provide rich information for these activities. Interestingly, visual and tactile information often exhibit significant relevance and complementarity ~\cite{apkarian1975comparison}. For individuals with visual impairments, a clearer mental reconstruction of an original visual image can be achieved by combining a blurred visual perception with tactile information~\cite{kappers2011human}.

While prior visuomotor learning often relies solely on vision ~\cite{pitz2023dextrous, lin2024learning, xu2023efficient, qi2023hand, lei2024unio, lei2025rl, yang2025novel}, recent works incorporating touch like M3L \cite{sferrazza2023power} and VTT \cite{chen2022visuo} typically use direct concatenation or simple patching. However, these approaches fail to exploit the inherent correspondence between visual and tactile information, resulting in suboptimal performance on challenging manipulation tasks. Moreover, many prior approaches overlook the complementary nature of visual and tactile modalities, which could undermine performance in self-occluded scenarios and  potentially limit their practical deployment. Given these limitations, we pose the question: \textit{how can we more effectively fuse visual and tactile information to enhance the performance of visuomotor algorithms}?

Drawing on prior research in human physiology regarding the processing of visuo-tactile information, we propose \textbf{Vi}sual \textbf{Ta}ctile \textbf{S}oft Fusion Contrastive Learning~(ViTaS), a novel visuo-tactile representation learning framework for visuomotor learning. Generally, ViTaS can be divided into two parts. Firstly, given the inherent relevance between visual and tactile modalities, we utilize contrastive learning to align the embeddings of visual data with their corresponding tactile information in the latent space. Notably, we employ \textit{soft fusion contrastive learning}, a novel method extending the RGB single-modality framework presented in CoCLR \cite{han2020self} to fuse features in alternating modalities. Second, inspired by humans' remarkable ability to reconstruct clear mental images from blurred visual input when supplemented with tactile  information, we integrate a conditional variational autoencoder (CVAE) \cite{sohn2015learning} that leverages the complementary nature of two modalities. The CVAE reconstructs visual observations from fused visuo-tactile embeddings, enforcing cross-modal consistency and enhancing the quality of multimodal representations.

To evaluate the performance of our algorithm, we conduct extensive experiments across both simulated and real-world domains with RL and IL paradigms. Our simulation benchmark comprises $12$ diverse manipulation tasks spanning $5$ distinct environments. To rigorously assess generalization capabilities, we extend our evaluation to $3$ auxiliary tasks featuring varied object geometries and environmental conditions, complemented by comprehensive ablation studies that dissect each component's contribution. Furthermore, to validate real-world applicability and handle more complex manipulation scenarios, we deploy ViTaS within IL paradigm for challenging real-world tasks, thereby establishing its practical viability beyond simulation environments. Our experimental results demonstrate that ViTaS achieves state-of-the-art performance compared with existing visuo-tactile learning baselines. 

In summary, our contributions are as follows:
\begin{itemize}
    \item We improve the traditional contrastive learning method and use it for the fusion of visual and tactile modalities.
    \item We propose ViTaS, a simple yet effective representation learning paradigm that can integrate visual and tactile inputs through soft fusion contrastive as well as CVAE, utilizing it to guide the training of visuomotor learning.
    \item We evaluate our algorithm on various tasks in both simulation and real-world environment, demonstrating state-of-the-art performance against various baselines.
\end{itemize}

\section{Related Work}
\label{sec:related_work}

\begin{figure*}[ht]
\begin{center}
\includegraphics[width=0.75\linewidth]{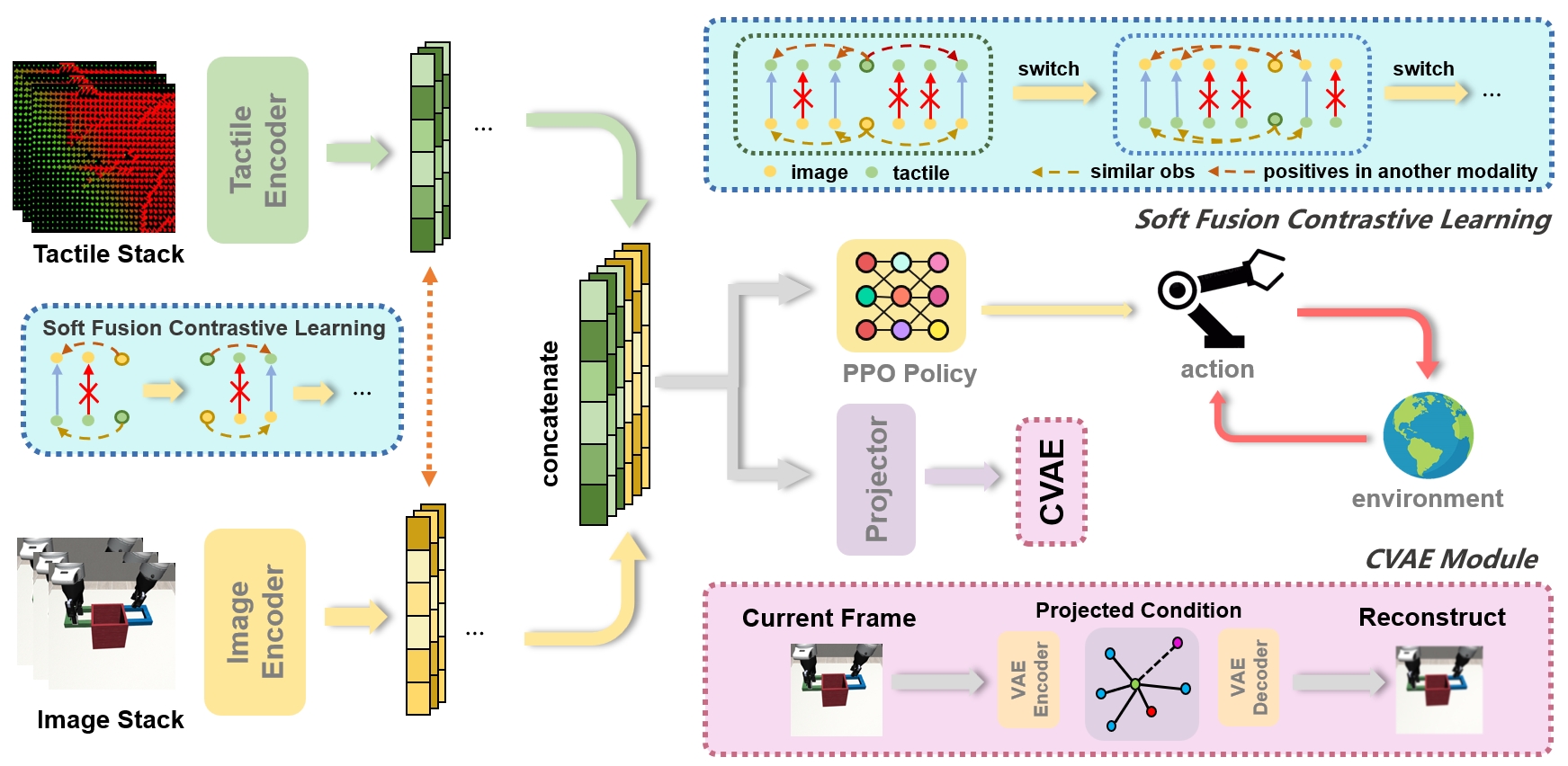}
\caption{\ex{\textbf{Method overview.} ViTaS takes vision and touch as inputs, which are then processed through separate CNN encoders. Encoded embeddings are utilized by \textit{soft fusion contrastive} approach, yielding fused feature representation for policy network. A CVAE-based reconstruction framework is also applied for \ex{cross-modal} integration.}}
\vspace{-7mm}  
\label{fig:overview}
\end{center}
\end{figure*}

\paragraph{Visuo-Tactile Representation Learning}
In recent years, numerous cross-modal representation learning methods have emerged, particularly those focused on visuo-tactile integration, as demonstrated by 
\cite{lee2020making, lee2019making, dave2024multimodal, yang2024binding, li2022see, yang2023sim, lin2024learning, xu2023efficient,li2022tata,lu2025h}. \ex{Among them}, VTT \cite{chen2022visuo} leverages a transformer architecture to integrate multiple modalities, introducing alignment and contact \ex{loss} to enhance performance.  M3L \cite{sferrazza2023power} proposes a jointly visuo-tactile training scheme using an MAE-based encoder trained through a reconstruction process.

Despite the success of these approaches in specific tasks, they often fail to fully exploit the correspondence between visual and tactile modalities, leading to suboptimal performance. In contrast, our method employs a simpler yet highly effective CNN-based encoder to improve the alignment and fusion of modalities, achieving superior performance.
\paragraph{Contrastive Learning}
Contrastive learning~\cite{he2020momentum, chen2021empirical, chen2020simple} has emerged as a prominent technique for representation learning. We intend to extend the contrastive learning paradigm to a visuo-tactile framework. Related examples include \cite{ContrastiveLearningMethodsforDRL}, \cite{dave2024multimodal}, \cite{yuan2021multimodal}, \cite{laskin2020curl}, \cite{wang2023contrastive}, \cite{han2020self}, \cite{yang2024binding}, \cite{ContrastiveLearningMethodsforDRL}, \cite{yuan2024learning}. 
\ex{Among the works most closely related to ours, MViTaC \cite{dave2024multimodal} proposes a visuo-tactile fusion approach based on contrastive pre-training.  UniTouch \cite{yang2024binding} uses tactile, vision, and text using contrastive learning to solve downstream tasks.}

However, simply doing instance discrimination tends to neglect some key information \cite{han2020self} since two resembling samples may be negatives for each other due to distinct timesteps. The phenomenon also pops up in the field of cross-modal contrastive learning. To alleviate the issue and better integrate different modalities, we refine the contrastive learning method, which is elaborated in \Cref{sec:Cross-modal Soft Contrastive Learning}.

\section{Method}

In this section, we elaborate \textbf{Vi}sual \textbf{Ta}ctile \textbf{S}oft Fusion Contrastive Learning~(ViTaS), an advanced visuo-tactile fusion framework tailored for visuomotor learning illustrated in \Cref{fig:overview}. To exploit the similarity between visual and tactile information, we propose \textit{soft fusion contrastive learning}~(\Cref{sec:Cross-modal Soft Contrastive Learning}), an extension of the conventional contrastive learning paradigm. Furthermore, to leverage their complementarity and handle the self-occluded scenarios, we integrate both modalities to guide the CVAE~(\Cref{sec:Conditional VAE Visual-Tactile Feature Integration}). Next, we elaborate these main designs with details.

\subsection{Soft Fusion Contrastive Learning}
\label{sec:Cross-modal Soft Contrastive Learning}

We define a trajectory as $\Gamma = \{o_i, t_i\}_{i=1}^{N}$, where $o_i$ and $t_i$ denote the visual and tactile observations at timestep $i$, respectively. The encoders $f_o(\cdot)$ and $f_t(\cdot)$ map these observations into their latent representations. For simplicity, we refer to $o_i$ and $t_i$ at the same timestep as \textit{dual} counterparts.

Inspired by CoCLR \cite{han2020self}, we introduce \textit{soft fusion contrastive learning}, a novel cross-modal contrastive paradigm designed to enhance multi-modal fusion. We refer to it as \textit{soft fusion contrastive} for brevity. The central intuition is that visual and tactile representations often exhibit strong structural alignment in non-occluded scenarios, making them particularly well-suited for contrastive learning.

In the first stage, given an image observation $o_i$, we retrieve its top-$K$ most similar images $o_{i_1}, \cdots, o_{i_K}$. The corresponding \textit{dual} tactile embeddings $t_{i_1}, \cdots, t_{i_K}$ are then designated as positive samples for $t_i$, while all other tactile maps are treated as negatives. With these positive and negative pairs, we apply a contrastive learning objective to train the encoder. Crucially, during this stage, only the tactile encoder is updated, while the visual encoder remains frozen. The resulting contrastive loss $\mathcal{L}_{\tt{CON},1,i}$ is formulated in Eq.~(1).
\begin{align}
    \left\{
\begin{array}{l}
\mathcal{L}_{\tt{CON}, 1, i} = - \log \dfrac{\mathcal{F}_{1, \mathcal{P}_{1i}}(i)}{\mathcal{F}_{1, \mathcal{P}_{1i}}(i) + \mathcal{F}_{1, \mathcal{N}_{1i}}(i)} \\
    \mathcal{F}_{1,S}(i) = \sum_{k\in S}{\exp(f_t(t_k) \cdot f_t(t_i) \ /\ \tau)} \\
\end{array}
    \right.
\end{align}


Formally, $\mathcal{P}_{1i}$ denotes the positive set of $t_i$, defined as the top-$K$ most similar visual samples formulated in Eq.~(2), while $\mathcal{N}_{1i}=S \setminus \mathcal{P}_{1i}$ denotes the negatives, where $S$ is the full sample set. The operator $\mathit{topKmax}_k(f_o(o_i) \oplus f_o(o_k))$ retrieves the top-$K$ most similar elements to the visual feature of $o_i$. The similarity between $x$ and $y$ is measured via cosine similarity, denoted as $x \oplus y$. $K$ is a hyperparameter set to $10$, with its ablation results presented in \Cref{AblationStudy}.
\begin{align}
\mathcal{P}_{1i}=\{j|(f_o(o_j) \oplus f_o(o_i))\in \mathit{topKmax}_k
    (f_o(o_i) \oplus f_o(o_k))\}
\end{align}

In the next stage, to avoid modality imbalance, we periodically swap the roles of $o_i$ and $t_i$, so that the visual encoder could also be updated. The corresponding objective and positive set are denoted as $\mathcal{L}_{\tt{CON},2,i}$ and $\mathcal{P}_{2i}$, formulated in Eq. (3) and Eq. (4). We also have $\mathcal{N}_{2i}=S \setminus \mathcal{P}_{2i}$.
\begin{align}
    \left\{
\begin{array}{l}
\mathcal{L}_{\tt{CON}, 2, i} = - \log \dfrac{\mathcal{F}_{2, \mathcal{P}_{2i}}(i)}{\mathcal{F}_{2, \mathcal{P}_{2i}}(i) + \mathcal{F}_{2, \mathcal{N}_{2i}}(i)} \\
    \mathcal{F}_{2,S}(i) = \sum_{k\in S}{\exp(f_o(o_k) \cdot f_o(o_i) \ /\ \tau)} \\
\end{array}
    \right.
\end{align}
\begin{align}
\mathcal{P}_{2i}=\{j|(f_t(t_j) \oplus f_t(t_i))\in \mathit{topKmax}_k
    (f_t(t_i) \oplus f_t(t_k))\}
\end{align}

To further stabilize training, we alternate between $\mathcal{L}_{\tt{CON},1,i}$ and $\mathcal{L}_{\tt{CON},2,i}$ using a switching schedule with period $T_{\tt{switch}}$. Specifically, we define a coefficient sequence $u_i= 1/2\times \left(1-(-1)^{\lceil i/T\rceil}\right)=[1,1,\cdots,1,0,0,\cdots,0,1,1,\cdots]$, yielding a binary mask that cycles between consecutive blocks of ones and zeros. Full procedure is illustrated in the aqua block of \Cref{fig:overview}. The overall contrastive objective is therefore:
\begin{align}
    \mathcal{L}_{\tt{CON}} = \sum_{i=1}^N(u_i \cdot \mathcal{L}_{\tt{CON}, 1, i}+(1-u_i)\cdot \mathcal{L}_{\tt{CON}, 2, i})
    \label{eq: con=1+2}
\end{align}


A natural counterargument is that soft fusion contrastive may be unnecessary, since temporally adjacent observations are already similar and could be used as positives. However, our ablation study (\Cref{AblationStudy}) demonstrates that this temporal heuristic is suboptimal compared to our approach, and we further provide an in-depth analysis of the reasons.

\subsection{Conditional VAE Visuo-Tactile Feature Integration}
\label{sec:Conditional VAE Visual-Tactile Feature Integration}
VAE-based methods are widely adopted for visuo-tactile integration~\cite{nair2018visual, bai2021variational}. In self-occluded scenarios, the agent should exploit the complementarity between visual and tactile modalities rather than relying solely on their alignment. Given the complementary information tactile map and image offer, our objective is to fully exploit the fusion of image and tactile embeddings, which could show superior performance to using the embeddings solely. To this end, we adopt CVAE to reconstruct observations from the fused embeddings \cite{CVAE-use}, thereby enforcing cross-modal consistency. A comprehensive depiction is presented in the pink block of \Cref{fig:overview}.

Specifically, we establish \textit{condition} on the concatenated visuo-tactile feature $c$ to reconstruct the current image frame $o_{\text{cur}}$. CVAE consists of an encoder $p_{\theta}(\cdot)$, decoder $q_{\psi}(\cdot)$, and visuo-tactile embedding projector $f_{\phi}(\cdot)$, which are parameterized by $\theta$, $\psi$ and $\phi$ separately. We use $z$ to represent the latent variables, and the reconstructed frame $\hat{o}_{\text{cur}}$ conditioned on visuo-tactile feature $c$ can be expressed as:
\begin{align}
\hat{\mathbf{o}}_{\text{cur}} = q_{\psi}(p_{\theta}(o_{\text{cur}}, f_{\phi}(c)), f_{\phi}(c))
\end{align}

In accordance with CVAE constraints, the target can be formulated as:
\begin{align}
\label{eq: VAE}
\mathcal{L}_{\tt{VAE}} = \mathbb{E} \left[ \| o_{\text{cur}} - \hat{o}_{\text{cur}} \|^2 \right] + D_{\text{KL}} \left( p_{\theta}(z|o_{\text{cur}}, c) \| \ \mathcal{N}(0, 1) \right)
\end{align}
During training, the CVAE encoder, decoder, projector, and the image and tactile encoders are jointly optimized. Only the image and tactile encoders are retained for inference.

With two representation loss, we reach the final objective:
\begin{align}
\mathcal{L} = \lambda\mathcal{L}_{\tt{CON}} + \mu\mathcal{L}_{\tt{VAE}} + \mathcal{L}_{\tt{policy}}
\end{align}
where $\lambda$ and $\mu$ are the coefficients to balance, and extra policy loss $\mathcal{L}_{\tt{policy}}$ depends on the RL or IL paradigms we adopt. In RL case policy loss is $\mathcal{L}_{\tt{PPO}}$, while $\mathcal{L}_{\tt{DP}}$ otherwise. The value of $\lambda, \mu$ is elaborated in \Cref{AblationStudy}.
\section{Experiments}

\label{sec:exp}
We evaluate our method on several contact-rich tasks in both simulation and real-world environments, in order to clarify the following questions: 

\begin{itemize}
  \item [(i)]
  \ex{Does ViTaS have the capability to solve complicated manipulation tasks~(e.g. dexterous hand rotation)?}
  \item [(ii)]
  \ex{How does ViTaS demonstrate generalization and robustness in tasks involving objects of various shapes, significant noise or different physical parameters?}
  \item [(iii)] 
  \ex{How does ViTaS perform in real-world settings?}
\end{itemize}

All three questions will be elaborated in the following parts. Moreover, ablation and qualitative studies are also conducted for the understanding of components in ViTaS.

\subsection{Simulation Environment Setup}
\subsubsection{Tasks}
\begin{figure}[h]
\vspace{-4mm}
\begin{center}
\includegraphics[width=1\linewidth]{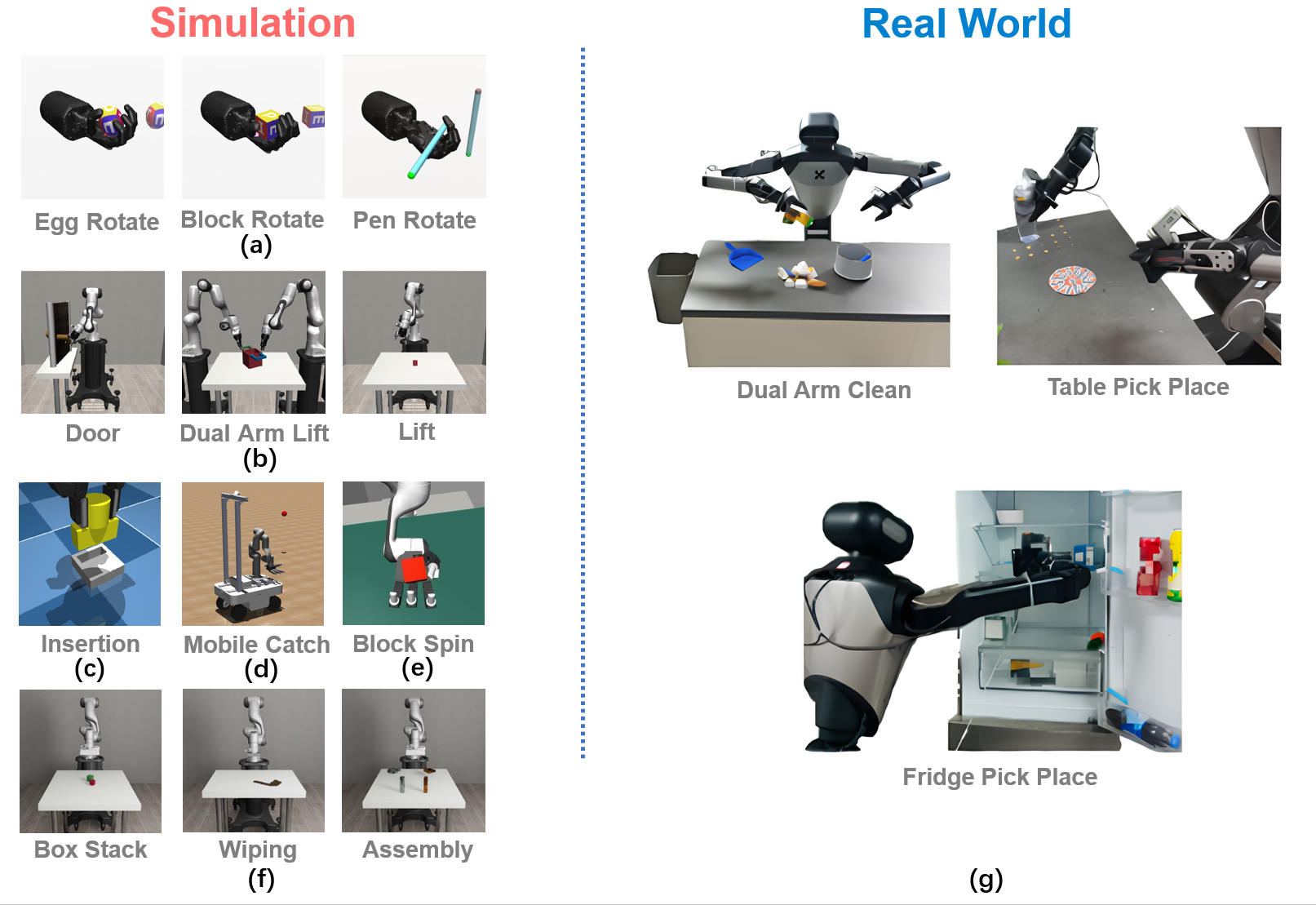}
\end{center}
\caption{\textbf{Tasks.} Our method is evaluated on $12$ simulation tasks and $3$ real-world tasks, with various embodiment types.}
\label{fig: tasks decription}
\vspace{-3mm}
\end{figure}
We evaluate on $12$ simulated tasks (\Cref{fig: tasks decription}) from various benchmarks, including: dexterous hand rotation (Gymnasium), contact-rich manipulation (Robosuite \cite{zhu2020robosuite}), Insertion \cite{sferrazza2023power}, Mobile Catching \cite{zhang2024catch}, and Block Spinning \cite{yuan2024learning}. Beyond these foundational experiments, we introduce a series of auxiliary tasks \ex{involving} altering object shapes in Lift or modifying physical parameters in Pen Rotation. The outcomes of (a)-(d), (f) environments are quantified in terms of success rate, and (e) is assessed based on training reward. (a)-(e) are trained with PPO, an effective reinforcement learning paradigm, while (f) with imitation learning is shown in \Cref{sec: sim imitation learning}.


\subsubsection{Tactile sensors}

It is crucial to integrate tactile sensors to obtain tactile data for ViTaS framework. For the $3$ in-hand rotation and Wiping tasks, we employ the built-in tactile modules. For Lift, Insertion, Door Opening, Box Stack and Assembly, we employ a parallel gripper equipped with a $32\times32\times3$ tactile sensor at the contact surfaces between the gripper and the object. \ex{Among the $3$ channels of the tactile map, channel $1$ and $2$ represent the normal force and channel $3$ denotes shear force, following \cite{taylor2022gelslim} and \cite{xu2023efficient}}. In the Mobile Catching and Block Spin tasks, we enhanced the dexterous hands with tactile sensors by integrating four $3\times 3\times 3$ sensors on each finger (located at the proximal, middle, distal, and tip segments) and one $3\times 3\times 3$ sensor on the palm. Sensors are \ex{zero-padded} to form a $32\times32\times3$ input for consistency.


\subsubsection{Comparison methods}
We compare ViTaS against $6$ visuo-tactile representation learning baselines:

\begin{itemize}
  \item
  M3L~\cite{sferrazza2023power}: A visuo-tactile fusion training algorithm utilizing the MAE encoder for PPO policy learning.
  \item
  VTT~\cite{chen2022visuo}: A visuo-tactile fusion training method rooted in the transformer architecture.
  \item
  PoE~\cite{lee2020making}: A VAE-like framework to fuse two modality.
  \item
  Concat~\cite{lee2019making}: A multi-modal fusion method with contrastive method.
  \item
  MViTac~\cite{dave2024multimodal}: A framework based on contrastive pre-training to fuse two modality, abbreviated as MVT.
  \item
  ConViTaC~\cite{wu2025convitac}: A visuo-tactile fusion method with contrastive learning, abbreviated as CVT.
\end{itemize}

\subsubsection{ViTaS with Imitation Learning}

\begin{figure}[h]
\vspace{-5mm}
\begin{center}
\includegraphics[width=0.8\linewidth]{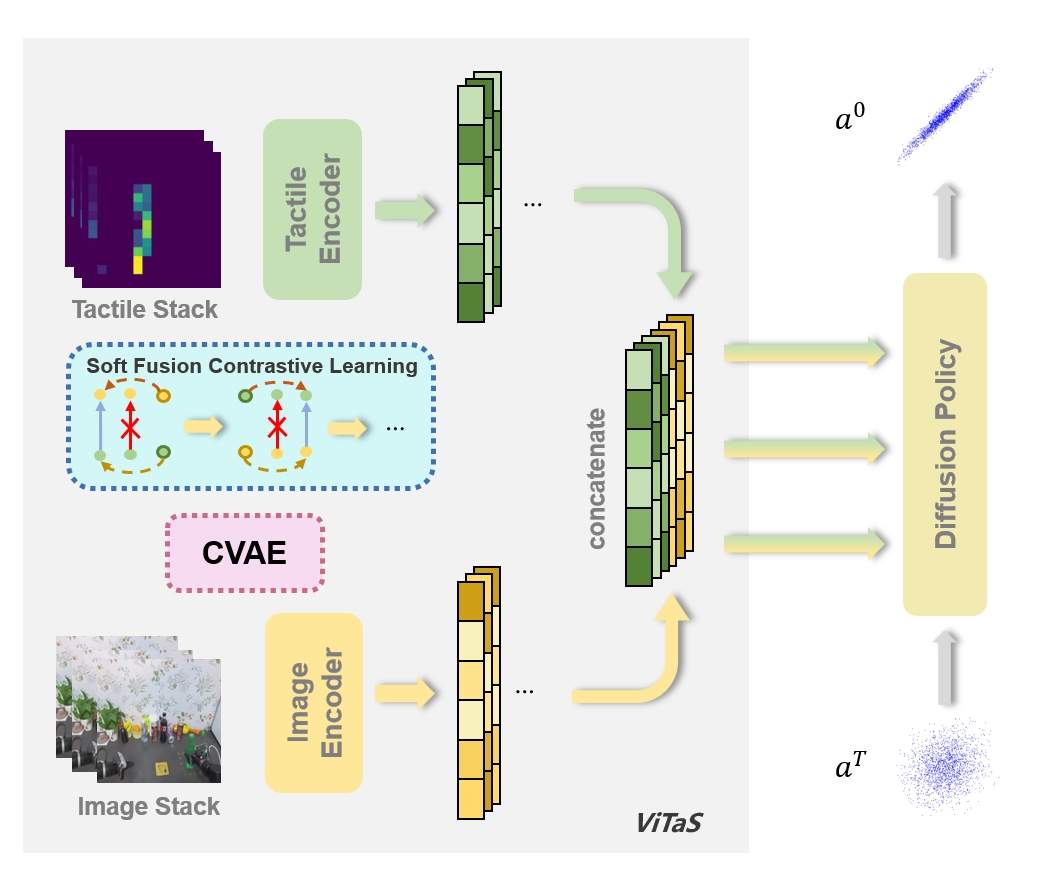}
\end{center}
\caption{\textbf{ViTaS with Imitation Learning.} The imitation learning paradigm is adopted to further test feature extraction ability of ViTaS in different settings, which we use for both simulation and real world tasks.}
\label{fig: vitas with dp}
\vspace{-3mm}
\end{figure}
\label{sec: sim imitation learning}

To further assess ViTaS’s capability in extracting and integrating multimodal features across diverse scenarios, we adopt Diffusion Policy (DP) \cite{chi2023diffusion}, a pioneering generative framework for robotic manipulation that formulates action prediction as a conditional denoising process, for both simulation and real-world evaluations. As illustrated in \Cref{fig: vitas with dp}, we replace default CNN or transformer encoders in original DP with those from ViTaS, while retaining the same training procedure for soft fusion contrastive and CVAE modules as in the reinforcement learning paradigm. For example, within the CVAE module, images are reconstructed from combined visuo-tactile embeddings. We evaluate ViTaS on $3$ simulation tasks, Box Stack, Wiping and Assembly, against original DP. To ensure a fair comparison, original DP receives inputs from $3$ cameras, whereas ViTaS relies solely on a single head camera and tactile sensors. This setting highlights that properly leveraging tactile features can compensate for reduced image, exhibiting robustness under limited visual input.


\subsection{Simulation Experiment Results}

\subsubsection{RL and IL results}
Our algorithm is compared against $4$ baseline methods across $9$ primitive tasks ((a)-(e) in \Cref{fig: tasks decription}) for reinforcement learning. We evaluate each algorithm in each environment $5$ times under different random seeds, and average the results when training $3\times 10^6$ timesteps to obtain the performance metrics. As for imitation learning, we train all methods across $3$ harder tasks ((f) in \Cref{fig: tasks decription}) for $10^3$ epochs, with $50$ collected expert demonstrations per task.


\renewcommand{\arraystretch}{1.2} 
\definecolor{tablecolor2}{RGB}{249, 230, 255}
\definecolor{tablecolor3}{RGB}{233, 255, 230}
\begin{table}[h] \caption{\textbf{{Performance.}} Each experiment repeats $5$ times. \colorbox{tablecolor3}{\text{Green}} for optimal results while \colorbox{tablecolor2}{\text{purple}} for suboptimal.}
\label{table1: generalization}
\centering
\renewcommand\tabcolsep{3.0pt}
\begin{footnotesize}
\begin{tabular}{ccccccccc}
\toprule[0.2mm]
\begin{tabular}[c]{@{}c@{}}Tasks / Methods\end{tabular}& \begin{tabular}[c]{@{}c@{}}{ViTaS} \end{tabular}  & \begin{tabular}[c]{@{}c@{}} MVT \end{tabular}& \begin{tabular}[c]{@{}c@{}} CVT \end{tabular}  & \begin{tabular}[c]{@{}c@{}} M3L \end{tabular}  & \begin{tabular}[c]{@{}c@{}} {PoE} \end{tabular}  & \begin{tabular}[c]{@{}c@{}} VTT \end{tabular}  & \begin{tabular}[c]{@{}c@{}} Concat\end{tabular}  \\  \hline

\begin{tabular}[c]{@{}c@{}}Insertion \end{tabular}       & \ddbf{98.2}\scriptsize & 64.3\scriptsize & \dd{83.9}\scriptsize & 72.1\scriptsize & 11.4 & 78.6\scriptsize & 19.3\scriptsize \\
\begin{tabular}[c]{@{}c@{}}Door\end{tabular}       & \ddbf{100.0}\scriptsize & \ddbf{100.0}\scriptsize & \ddbf{100.0}\scriptsize  & \ddbf{100.0}\scriptsize & 98.2\scriptsize &\dd{99.8}\scriptsize&\ddbf{100.0}  \\
\begin{tabular}[c]{@{}c@{}}Lift\end{tabular}         & \dd{97.5}\scriptsize  & \ddbf{97.9}\scriptsize & 70.2\scriptsize  & 20.6\scriptsize & 71.9\scriptsize &70.4\scriptsize&76.7 \\
\begin{tabular}[c]{@{}c@{}}Pen Rotate\end{tabular}      & \ddbf{99.2}\scriptsize & 77.1\scriptsize & \dd{79.6}\scriptsize & 73.1\scriptsize & 0.3 & 0.7\scriptsize & 2.9\scriptsize\\
\begin{tabular}[c]{@{}c@{}}Dual Arm Lift\end{tabular}     & \ddbf{100.0}\scriptsize & 87.5\scriptsize & 90.8\scriptsize   & 88.2\scriptsize & \dd{92.6}\scriptsize &77.1\scriptsize&76.8 \\
\begin{tabular}[c]{@{}c@{}}Mobile Catch\end{tabular}   & \ddbf{64.9}\scriptsize& 44.2\scriptsize & \dd{64.2}\scriptsize   & 15.8\scriptsize & 0.4\scriptsize &53.3\scriptsize&0.6\\
\begin{tabular}[c]{@{}c@{}}Egg Rotate\end{tabular}      & \ddbf{85.7}\scriptsize & \dd{71.6}\scriptsize & 58.1\scriptsize& 4.2\scriptsize & 0.9 &0.5\scriptsize &0.7\scriptsize  \\
\begin{tabular}[c]{@{}c@{}}Block Rotate\end{tabular}        & \ddbf{93.3}\scriptsize & 69.1\scriptsize & \dd{70.4}\scriptsize  &11.6\scriptsize & 0.8\scriptsize &1.3\scriptsize&4.4  \\
\begin{tabular}[c]{@{}c@{}}Block Spin\end{tabular}        & \ddbf{70.5}\scriptsize & 35.7\scriptsize & \dd{43.4}\scriptsize  & 30.8\scriptsize & 20.6\scriptsize &0.9\scriptsize&15.7\\
\begin{tabular}[c]{@{}c@{}}Insertion Noisy \end{tabular}      & \ddbf{89.2}\scriptsize& 51.6\scriptsize & \dd{70.5}\scriptsize &47.3\scriptsize & 20.7 & 63.4\scriptsize & 26.9\scriptsize  \\
\begin{tabular}[c]{@{}c@{}}Lift w/ Cap\end{tabular}       & \ddbf{99.6}\scriptsize & 67.8\scriptsize & 49.9\scriptsize  & 54.2\scriptsize & 58.1\scriptsize &54.7\scriptsize&\dd{87.5}  \\
\begin{tabular}[c]{@{}c@{}}Lift w/ Can\end{tabular}       & \ddbf{97.8}\scriptsize & \dd{76.7}\scriptsize & 55.9\scriptsize & 41.3\scriptsize & 52.6\scriptsize &69.4\scriptsize&75.8 \\
\hline
\begin{tabular}[c]{@{}c@{}}Average\end{tabular}    & \ddbf{91.4}\scriptsize & 70.3\scriptsize & \dd{71.5}\scriptsize  & $46.6$\scriptsize & ${35.7}$\scriptsize &47.5\scriptsize&40.6 \\

\midrule[0.3mm]
\end{tabular}

\end{footnotesize}
\vspace{-5pt}
\label{tab: simulation results}
\end{table}



\begin{table}[htb]
    \vspace{-4mm}
    \centering
    \caption{\textbf{Simulation imitation learning results.} Each experiment repeats $3$ times with different random seeds.}
    \begin{tabular}{ccccc}
        \toprule[0.4mm]
        Methods / Tasks & \makecell{Box Stack} & \makecell{Wiping} & \makecell{Assembly} & Average\\
        \midrule[0.2mm]
        DP w/ ViTaS & \cellcolor{tablecolor3}$\mathbf{53.3}$ & \cellcolor{tablecolor3}$\mathbf{71.7}$ & \cellcolor{tablecolor3}$\mathbf{56.3}$ & \cellcolor{tablecolor3}$\mathbf{60.4}$\\
        DP w/ CNN & 29.7 & 57.0 & 28.0 & 38.2\\
        DP w/ Transformer & 30.3 & 45.0 & 25.0 & 33.4\\
        \hline
    \end{tabular}
    \label{table:sim-il-results}
\end{table}
\begin{table}[htb]
    \centering
    \vspace{-5mm}
    \caption{\textbf{Generalization ability.}}
    \begin{tabular}{ccccc}
        \toprule[0.4mm]
        Tasks / Methods & ViTaS & MVT& CVT& M3L \\
        \midrule[0.2mm]
        Pen Rotate w/ Fixed Target & \cellcolor{tablecolor3}$\mathbf{99.2}$ & $77.1$  & $79.6$ & $73.1$\\
        Pen Rotate w/ Random Target & \cellcolor{tablecolor3}$\mathbf{78.4}$ & $47.5$  & $51.3$ & $42.7$\\
        \bottomrule
    \end{tabular}
    \vspace{-3mm}
    \label{table: generalization ability}
\end{table}

As the results shown in \Cref{tab: simulation results}, several baselines show excellent performance in simple tasks like Door Open and Insertion. However, for tough tasks like Egg Rotate and Block Rotate, which require methods to incorporate visual and tactile information jointly, few baselines can solve it within a limited horizon, while ViTaS maintains its performance. Moreover, as shown in \Cref{table:sim-il-results}, ViTaS shows superior performance against DP in all $3$ tasks, showing the huge difference touch could make in manipulation tasks. \textbf{This underscores its exceptional capability to extract features and solve complex tasks, clarifying question (i)}. 

\subsubsection{Generalization and robustness results}

In order to assess the generalization capability of our approach, we introduce auxiliary variants of the previously mentioned Lift and Pen Rotate tasks. the object shape in Lift task is modified from a cube to cylinder and capsule in both training and testing phases, allowing us to evaluate the method's resilience to changes in object geometry. As for the Pen Rotate task, we randomize the target angle within a large range to test the generalization of methods. Furthermore, to test the robustness, \ex{we also add Gaussian noise with $0.3$ noise level in Insertion task. The intensity of Gaussian noise of different noise levels could refer to \Cref{fig:recon}.} All other settings are in alignment with the preceding $9$ tasks.

As illustrated in \Cref{tab: simulation results}, when the object shape is changed, every baseline model experiences a performance drop \ex{when training $3\times 10^6$ timesteps}, indicating sensitivity to these alterations. ViTaS, however, exhibits a negligible decrease, demonstrating its resilience to variations in object geometry. Moreover, as the results shown in \Cref{table: generalization ability}, ViTaS demonstrates superior generalization ability to other baselines, given the performance in the randomized target position of Pen Rotate task. \textbf{The results across all auxiliary tasks provide a clear answer to question (ii).}

\subsection{Real-World Experiment}
\vspace{-5mm}  
\begin{figure}[h]
\begin{center}
\includegraphics[width=1.0\linewidth]{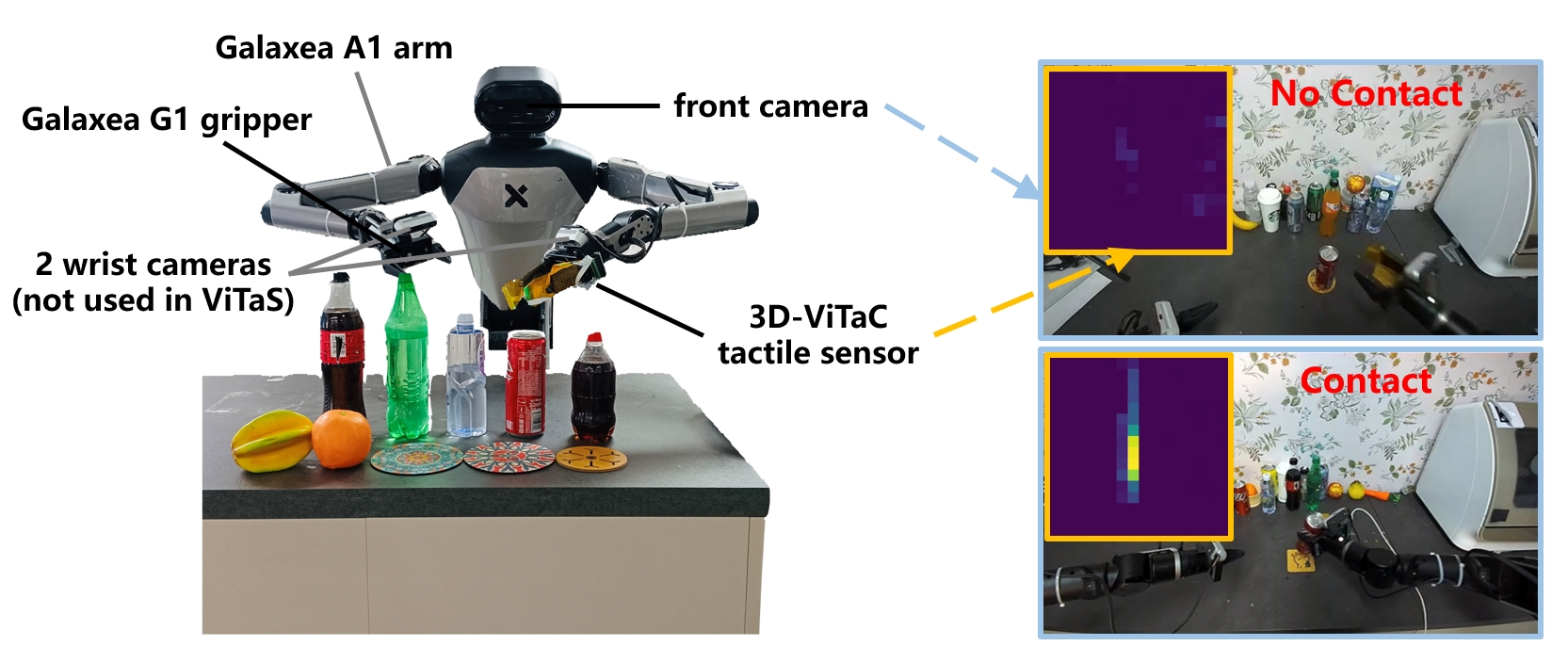}
\end{center}
\vspace{-4mm}  
\caption{\textbf{Real-World Robot Setting.}}
\label{fig: robot_overall}
\end{figure}
\vspace{-2mm}  


\subsubsection{Tasks}
To better understand the overall performance of ViTaS, we develop $3$ real-world experiments to show the effectiveness, shown in Figure \ref{fig: tasks decription}~(g): ~(1). Dual Arm Clean~(DAC). The robot has to sweep a small amount of rubbish ~(e.g. a piece of paper ball) to the trash can.~(2). Table Pick Place~(TPP). The robot has to move the bottles or cans to the coaster. This task has two settings: one~(TPP-1) uses a single type of bottle and the other~(TPP-3) uses three types.~(3). Fridge Pick Place~(FPP). The robot has to move the bottles from third level to the second of refrigerator.

\subsubsection{Experiment setup}
The overall working space is shown in Fig \ref{fig: robot_overall}. We use Galaxea-R1 Humanoid Robot for manipulation, with tactile sensors attached to the end effector.
\begin{itemize}
    \item Camera: we use multiple cameras to obtain RGB visual information. \textbf{Only} the head camera of Galaxea-R1~(Zed 2) is used in all ViTaS experiments, while $2$ wrist cameras~(RealSense D435i) are needed for better performance in some \textit{baseline} experiments.
    \item Tactile sensors: We use 3D-ViTaC \cite{huang20243d} producing real-time $16\times 16\times 1$ discrete 1D haptic maps for tactile information. The tactile sensors are attached to the gripper, obtaining tactile maps during data collecting and inference. A snapshot shown in \Cref{fig: robot_overall} demonstrates that when grippers contact objects, corresponding position of tactile maps have numerical changes (displayed in light colors) compared to others.
    \item Data collection: We collect $100$ real-world expert trajectories for each task, using Meta Quest 3 to teleoperate.
\end{itemize}
\subsubsection{ViTaS in real world}
As mentioned in \Cref{sec: sim imitation learning}, we also adopt imitation learning paradigm for real-world experiments, where the pipeline is shown in \Cref{fig: vitas with dp}. It is noteworthy that the real-world setting differs slightly from the simulation environment, requiring minor adaptations to achieve better empirical performance. To accommodate the 1D haptic map produced by our tactile sensor, the CNN tactile encoder is adjusted accordingly from $32\times32\times3$ to $16\times16\times1$. Both ViTaS and DP are trained with real data for $10^3$ epochs, and are then used for real-world R1 humanoid to calculate success rate in manipulation tasks. Since we use real data for training, there is no sim-to-real issues for ViTaS.


\subsubsection{Comparing method}
In resemblance with the simulation experiment settings, we compare ViTaS with DP in the real world. Given the CNN encoder shows better results in \Cref{table:sim-il-results}, we use learning-from-scratch CNN encoder for DP to assess the performance. Notably, ViTaS uses only the \textit{head} camera and tactile sensor as input, whereas DP leverages all three cameras (\textit{head, left, right}), thus receiving richer visual information. This comparison highlights that ViTaS achieves superior performance despite having access to less sensory input, further demonstrating its effectiveness.

\subsubsection{Results}
We have done Table Pick Place and Fridge Pick Place with $25$ repetitions, with the target position shifted slightly each time. We also complete Dual Arm Clean with $10$ repetitions, placing only \textit{one} piece of litter during benchmarking. The results are shown in \cref{table:real-world-results}.
\begin{table}[htb]
    \centering
    \vspace{-2mm}
    \caption{\textbf{Real-world experiment results.}}
    \begin{tabular}{cccccc}
        \toprule[0.4mm]
        Method / Tasks & \makecell{DAC} & \makecell{TPP-1} & \makecell{TPP-3} & \makecell{FPP} & Average\\
        \midrule[0.2mm]
        ViTaS & \cellcolor{tablecolor3}$\mathbf{30.0}$ & \cellcolor{tablecolor3}$\mathbf{42.0}$ & \cellcolor{tablecolor3}$\mathbf{36.0}$ & \cellcolor{tablecolor3}$\mathbf{76.0}$ & \cellcolor{tablecolor3}$\mathbf{46.0}$\\
        DP & 20.0 & 36.0 & 24.0 & 40.0 & $30.0$\\
        \hline
    \end{tabular}
    \vspace{-4pt}
    \label{table:real-world-results}
\end{table}

ViTaS consistently outperforms DP across $3$ real-world tasks, achieving an average success rate improvement of 16\%, despite relying on reduced camera input. Additionally, a comparative analysis of tasks TPP-1 and TPP-3 demonstrates ViTaS's superior generalization, with a smaller performance drop between tasks compared to DP. Notably, occlusions induced by the robotic arm in the head camera's field of view can impair visual perception. Nevertheless, ViTaS maintains robust performance, suggesting that tactile sensing effectively compensates for these visual limitations. This resilience highlights the efficacy of the CVAE mechanism in handling self-occlusion scenarios, underscoring ViTaS's ability to integrate both modalities seamlessly. Given the better performance in real-world settings, \textbf{we reach the answer to question (iii).}

\subsection{Ablation Study}
\label{AblationStudy}

To verify the fidelity of each component in ViTaS, we conduct extensive ablation experiments, showing the necessity of different designs. The general ablation results are presented in \Cref{tab: ablation study}, where we use abbreviations of experiments in the first row, corresponding to \textbf{V}iTaS, w/o. \textbf{TA}ctile information, w/ \textbf{U}nified Encoder, w/o. \textbf{S}oft fusion contrastive module, w/o. \textbf{C}VAE, w/ \textbf{T}ime \textbf{C}ontrastive, $K=1$, $K=20$ and $K=50$. The results are shown in columns V, TA, U, S, C, TC, K1, K20, K50, respectively. Detailed analysis of each experiment is clarified in the following sections.

\renewcommand{\arraystretch}{1.2} 
\definecolor{tablecolor2}{RGB}{249, 230, 255}
\definecolor{tablecolor3}{RGB}{233, 255, 230}
\begin{table}[h] \caption{\textbf{\ex{Ablation study.}} Each experiment repeats $5$ times. \colorbox{tablecolor3}{\text{Green}} for optimal results while \colorbox{tablecolor2}{\text{purple}} for suboptimal.}
\label{table1: generalization}
\centering
\renewcommand\tabcolsep{3.0pt}
\begin{footnotesize}
\begin{tabular}{cccccccccc}
\toprule[0.2mm]
\begin{tabular}[c]{@{}c@{}}Tasks\end{tabular} & \begin{tabular}[c]{@{}c@{}}V \end{tabular} & \begin{tabular}[c]{@{}c@{}}\ex{TA} \end{tabular}   & \begin{tabular}[c]{@{}c@{}} U \end{tabular}  & \begin{tabular}[c]{@{}c@{}} \ex{S} \end{tabular} & \begin{tabular}[c]{@{}c@{}} \ex{C} \end{tabular} & \begin{tabular}[c]{@{}c@{}} TC \end{tabular}  & \begin{tabular}[c]{@{}c@{}} K1 \end{tabular} & \begin{tabular}[c]{@{}c@{}} K20 \end{tabular}  & \begin{tabular}[c]{@{}c@{}} K50 \end{tabular}  \\  \hline

\begin{tabular}[c]{@{}c@{}}Insertion \end{tabular}      & \ddbf{99.2}    & $88.1$\scriptsize & $61.6$\scriptsize & \dd{90.3}& $71.1$\scriptsize & $75.2$\scriptsize & $83.3$\scriptsize & $85.1$\scriptsize& $78.7$\scriptsize \\
\begin{tabular}[c]{@{}c@{}}Block Rotate\end{tabular}    & \ddbf{92.7}        & $67.7$\scriptsize  & $18.4$\scriptsize & ${67.7}$\scriptsize &$70.2$\scriptsize&$79.5$\scriptsize&\dd{88.0}&$77.9$\scriptsize &$70.1$\scriptsize   \\
\begin{tabular}[c]{@{}c@{}}Egg Rotate\end{tabular}    & \ddbf{85.3}        & $24.3$\scriptsize  & $3.3$\scriptsize & ${6.5}$\scriptsize &$48.4$\scriptsize&$57.7$\scriptsize&\dd{65.2}&$41.2$\scriptsize &$3.6$\scriptsize   \\
\hline
\begin{tabular}[c]{@{}c@{}}Average\end{tabular}    & \ddbf{92.5}       & $60.9$\scriptsize  & $27.1$\scriptsize & ${54.7}$\scriptsize &$63.2$\scriptsize&$70.6$\scriptsize&\dd{78.8}&$68.1$\scriptsize&$67.3$\scriptsize   \\

\midrule[0.3mm]
\end{tabular}

\end{footnotesize}
\vspace{-5pt}
\label{tab: ablation study}
\end{table}


\textbf{Is tactile information crucial?} We conduct $2$ main experiments in this part. First we eliminate the tactile information, retaining only the visual data, and solely utilize the image encoder, while handling the corresponding tactile information through zero-padding. Additionally, the M3L workflow \cite{sferrazza2023power} employs a unified MAE encoder across both modalities which overlooks their inherent differences and may result in less discriminative features and reduced effectiveness. To further demonstrate that tactile maps provide complementary information requiring separate encoders, we also design an experiment where visual and tactile inputs are directly concatenated and processed by a shared encoder.


The TA-column in \Cref{tab: ablation study} show that when ablating tactile information, the success rate drops $32$\% on average. Thus, we prove that tactile information is crucial in manipulation tasks. Using a unified encoder, however, is not a good choice either, given the poor performance in the U-column in \Cref{tab: ablation study}, especially for the  $2$ rotation tasks. We attribute this result to the fact that unified encoder integrate both modalities roughly, neglecting the inherent discrepancy of visual and tactile maps which separate encoders could alleviate. 

\textbf{How much do CVAE and soft fusion contrastive contribute to ViTaS?} In order to clarify the effectiveness of each component, we remove the CVAE and soft fusion contrastive separately, conducting experiments on the same benchmarks. Results of ViTaS without CVAE or soft fusion contrastive \ex{are} shown in S and C-column of \Cref{tab: ablation study}. The performance drops heavily without CVAE or soft fusion contrastive learning, showing the necessity of two main designs.

\renewcommand{\arraystretch}{1.2} 
\definecolor{tablecolor2}{RGB}{249, 230, 255}
\definecolor{tablecolor3}{RGB}{233, 255, 230}

\textbf{$K$ in soft fusion contrastive learning}. We explore the impact of varying $K$, for instance, setting it to $1$, $10$~(ours), $20$ and $50$, to observe how the results are affected. It is noteworthy that image and tactile at the same timestep are the only positives for each other when $K=1$, adopting the same process as conventional cross-modal contrastive learning. Therefore, by comparing results between ours and $K=1$, we can also clarify whether soft fusion contrastive could outperform conventional contrastive learning method.

The last $3$ columns of \Cref{tab: ablation study} show the effectiveness of different $K$ in ViTaS. The results when $K=1$ show that though conventional contrastive learning can achieve relatively excellent performance, it still has gap with ViTaS~(i.e. $K=10$), while too large $K$ also causes performance drops.

\textbf{Soft fusion contrastive vs. time contrastive}. To verify the effectiveness of soft contrastive in another perspective, we carry out experiments utilizing an alternative contrastive approach, namely \textit{time contrastive}, to highlight the indispensable role of cross-modal soft fusion contrastive learning. Neighboring frames~(i.e. a fixed number of preceding and succeeding frames) are treated as positives in this method, while distant frames serve as negatives, echoing with TCN \cite{sermanet2018time}. The motivation behind the ablation lies in that, as mentioned in \Cref{sec:Cross-modal Soft Contrastive Learning}, despite frames within close time intervals often appearing to be similar, it is crucial during the contrastive learning process to identify the $K$ most analogous frames, which may not necessarily be temporally adjacent. This distinction underscores the importance of going beyond mere time contrastive. As shown in TC-column in \cref{tab: ablation study}, soft fusion contrastive learning is necessary as it outperforms time contrastive.

\textbf{Weight ablation in learning objective.} In Eq. (8), $\lambda, \mu$ indicate the weights of soft fusion contrastive and CVAE modules, and we initially set $\lambda=1, \mu=0.1$. To confirm that these parameters are indeed optimal, we conduct an ablation study by systematically varying the coefficients and evaluating the performance of our method on two particularly challenging tasks. As shown in \Cref{tab: coefficient ablation}, the performance when $\lambda=1, \mu=0.1$ is superior to other settings, showing the rationale of our chosen weights.
\begin{table}[h]
\vspace{-2mm} \caption{\ex{\textbf{Ablation on $\lambda, \mu$ in learning objective.}}}
\label{tab: coefficient ablation}
\centering
\renewcommand\tabcolsep{3.0pt}
\begin{footnotesize}
\begin{tabular}{ccccccc}
\toprule[0.2mm]
\begin{tabular}[c]{@{}c@{}}$\mu~(\lambda=1)$\end{tabular} & \begin{tabular}[c]{@{}c@{}}0.1~(ViTaS) \end{tabular} & \begin{tabular}[c]{@{}c@{}}0.01 \end{tabular}   & \begin{tabular}[c]{@{}c@{}} 1 \end{tabular}  & \begin{tabular}[c]{@{}c@{}} 10 \end{tabular}  & \begin{tabular}[c]{@{}c@{}} 0.5 \end{tabular}  & \begin{tabular}[c]{@{}c@{}} 0.05 \end{tabular}   \\  \hline

\begin{tabular}[c]{@{}c@{}}Egg Rotate \end{tabular}      & \ddbf{85.7}    & $42.5$\scriptsize & $67.7$\scriptsize & $58.4$\scriptsize & $68.6$\scriptsize & \dd{77.8} \\
\begin{tabular}[c]{@{}c@{}}Block Rotate\end{tabular}    & \ddbf{93.3}      & $60.3$\scriptsize  & $74.1$\scriptsize & $48.9$\scriptsize &\dd{80.2} &$76.5$\scriptsize   \\
\hline
\begin{tabular}[c]{@{}c@{}}$\lambda~(\mu=0.1)$\end{tabular} & \begin{tabular}[c]{@{}c@{}}1~(ViTaS) \end{tabular} & \begin{tabular}[c]{@{}c@{}}0.1 \end{tabular}   & \begin{tabular}[c]{@{}c@{}} 10 \end{tabular}  & \begin{tabular}[c]{@{}c@{}} 100 \end{tabular}  & \begin{tabular}[c]{@{}c@{}} 5 \end{tabular}  & \begin{tabular}[c]{@{}c@{}} 0.5 \end{tabular}  \\  \hline

\begin{tabular}[c]{@{}c@{}}Egg Rotate \end{tabular}      & \ddbf{85.7}    & $57.9$\scriptsize & $55.2$\scriptsize & $41.3$\scriptsize & $69.1$\scriptsize & \dd{78.8} \\
\begin{tabular}[c]{@{}c@{}}Block Rotate\end{tabular}    & \ddbf{93.3}      & $74.0$\scriptsize  & $66.9$\scriptsize & $43.2$\scriptsize & $79.8$ &\dd{85.1}\scriptsize   \\

\midrule[0.3mm]
\end{tabular}

\end{footnotesize}
\vspace{-5pt}
\label{tab: coef cvae}
\end{table}





In conclusion, our ablation study delves deep into our algorithm to analyze the effectiveness of each component. The results prove that tactile information, soft fusion contrastive learning and CVAE are of high importance, while soft fusion contrastive performs better than other contrastive methods like conventional contrastive learning and time contrastive.
\subsection{Qualitative Analysis}
To demonstrate the impact of the ViTaS, we use weights in the Egg Rotate task to reconstruct images from pure Gaussian noise conditioned on visuo-tactile embeddings. We compare performance under different noise levels in the observation space (visual and tactile) against the token-based MAE in M3L. Moreover, to validate CVAE’s role in cross-modal complementarity, we ablate the two main designs respectively by blocking gradients of soft fusion contrastive or CVAE propagating back to the image and tactile encoders, while keeping the CVAE encoder and decoder trainable. We then reconstruct occluded images given concatenated visuo-tactile features as condition from Gaussian noise. 

\begin{figure}[h]
    \begin{center}
    \includegraphics[width=0.9\linewidth]{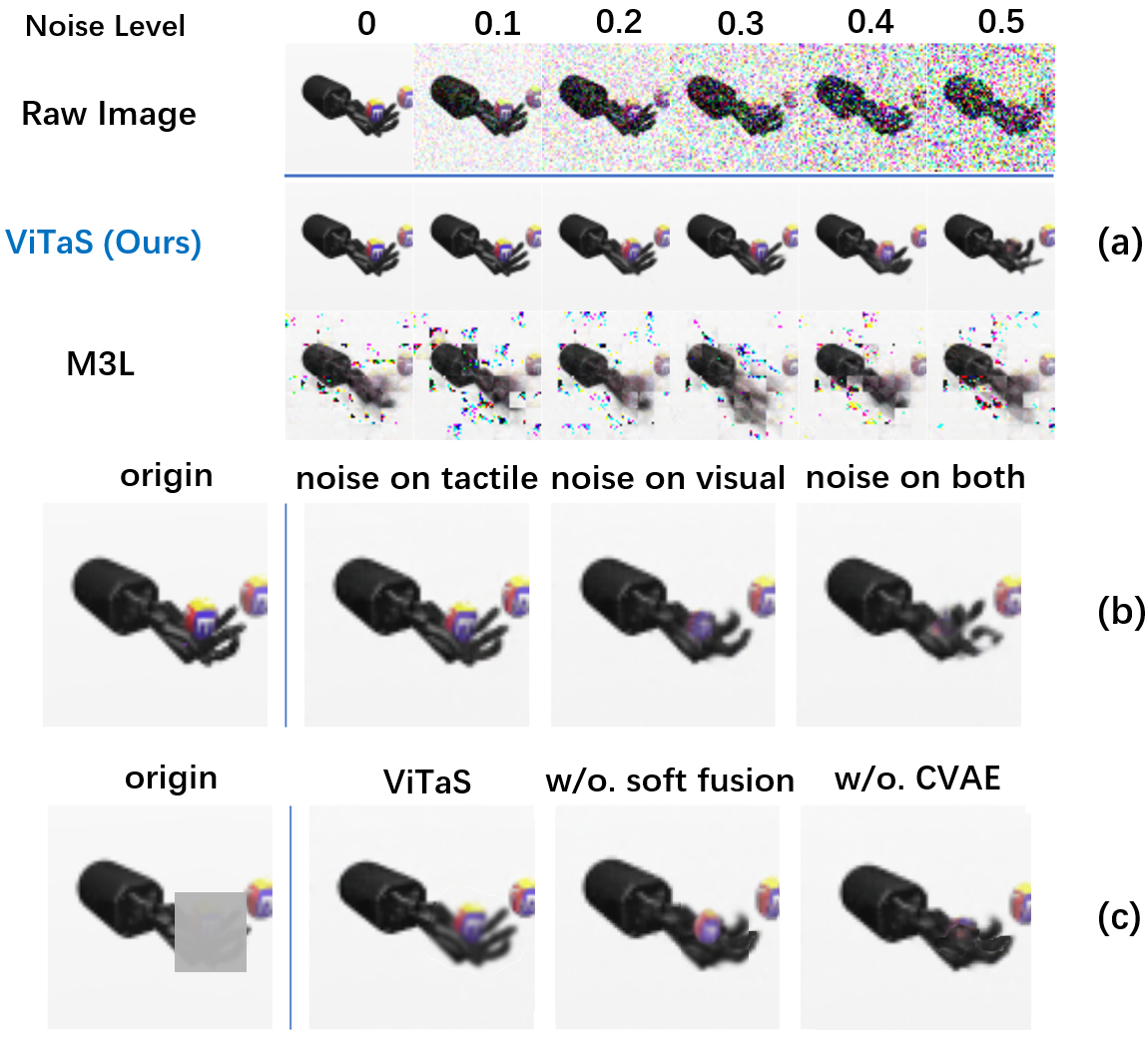}
    \end{center}
    \caption{\textbf{Reconstruction visualization.}~(a) compares the reconstruction quality of ViTaS and M3L under varying noise levels. (b) presents ViTaS reconstructions under heavy noise applied to different modalities. (c) illustrates reconstruction performance under masking when ablating key designs.}
    \label{fig:recon}
    \vspace{-5mm}
\end{figure}

As illustrated in \Cref{fig:recon}~(a), the results indicate that ViTaS surpasses the token-based MAE in reconstructing critical interaction details, such as the egg's location, which is vital for the task. Our method also maintains robust under higher level of noise, underscoring the high quality of the visuo-tactile embeddings used as conditions. In \Cref{fig:recon}~(b), applying heavy noise (noise level $0.5$) to one modality yields better reconstructions than adding noise to both, demonstrating visuo-tactile complementarity. \Cref{fig:recon}~(c) shows that even with core image regions masked, ViTaS reliably reconstructs observations when guided by tactile and masked visual features, validating CVAE’s role in handling self-occlusion and complementarity between two modalities.
\section{Conclusion and Limitations}

In general, we present ViTaS, a concise yet effective visuo-tactile fusion framework. Analogous to human physiology, it extends visual and tactile perception to RL and IL, achieving strong results in both simulation and real-world experiments. Specifically, \textit{soft fusion contrastive learning} extracts key features across modalities, while a CVAE module exploits their complementarity. Real-world experiments confirm ViTaS’s effectiveness, and ablation study plus qualitative analysis highlight the necessity of each component.

Despite its success, ViTaS faces $2$ main limitations. The first is that due to the physical capability and the bottleneck of RL in high-dimensional cases, some high dynamic accurate manipulation \cite{yuan2023rl} like pen spinning in real world remains challenging. The other is that the potential of visuo-tactile sensing in deformable object manipulation warrants extra exploration. In the future, we will further explore the ability of fused visuo-tactile features in more complex scenarios, with advanced simulation and real-world platforms.

\bibliographystyle{IEEEtran}
\bibliography{myrefs}

\vfill

\end{document}